# ShrimpXNet: A Transfer Learning Framework for Shrimp Disease Classification with Augmented Regularization, Adversarial Training, and Explainable AI

Israk Hasan Jone
Department of Computer Science and Engineering
Daffodil International University
hasan15-4119@diu.edu.bd

D.M. Rafiun Bin Masud
Department of Computer Science and Engineering
East West University
dmrafiun@gmail.com

Promit Sarker
Department of Computer Science and Engineering
North South University
promitsrkr@gmail.com

Sayed Fuad Al Labib
Department of Computer Science and Engineering
East West University
alfuad1007@gmail.com

Nazmul Islam
Department of Information Technology
University of Information Technology & Sciences
1914555005@uits.edu.bd

Farhad Billah
Department of Computer Science and Engineering
East West University
farhadbillah2020@gmail.com

*Abstract:* Shrimp is one of the most widely consumed aquatic species globally, valued for both its nutritional content and economic importance. Shrimp farming represents a significant source of income in many regions; however, like other forms of aquaculture, it is severely impacted by disease outbreaks. These diseases pose a major challenge to sustainable shrimp production. To address this issue, automated disease classification methods can offer timely and accurate detection. This research proposes a deep learning-based approach for the automated classification of shrimp diseases. A dataset comprising 1,149 images across four disease classes was utilized. Six pretrained deep learning models ,ResNet50, EfficientNet, DenseNet201, MobileNet, ConvNeXt-Tiny, and Xception were deployed and evaluated for performance. The images background was removed, followed by standardized preprocessing through the Keras image pipeline. Fast Gradient Sign Method (FGSM) was used for enhancing the model robustness through adversarial training. While advanced augmentation strategies, including CutMix and MixUp, were implemented to mitigate overfitting and improve generalization. To support interpretability, and to visualize regions of model attention, post-hoc explanation methods such as Grad-CAM, Grad-CAM++, and XGrad-CAM were applied. Exploratory results demonstrated that ConvNeXt-Tiny achieved the highest performance, attaining a 96.88% accuracy on the test dataset. After 1000 iterations, the 99% confidence interval for the model is [0.953,0.971]

*Keywords:* Shrimp Disease, Deep Learning, Transfer Learning, Adversarial Training, Augmented Regularization, Explainable AI

*I. Introduction*

Shrimp farming constitutes a vital economic pillar for Bangladesh, ranking among the world's top producers and exporters [1]. The industry sustains millions of coastal livelihoods while contributing approximately 5% to national export earnings, with critical markets in the United States, EU, and Japan. Beyond financial value, shrimp provides essential nutrition through high-quality protein, omega-3 fatty acids, and vital minerals [2]. However, disease outbreaks - particularly White Spot Syndrome Virus (WSSV) - cause catastrophic production losses exceeding 80% in affected ponds. These outbreaks directly undermine Bangladesh's export competitiveness, as evidenced by historical trade disruptions in Thailand and Vietnam. Traditional diagnostic methods fail to provide scalable, early detection across the country's fragmented smallholder farms, creating urgent need for automated solutions to safeguard production value.

Deep learning models have been proven to be an extremely effective tool for this paper. Various deep learning models have been successfully utilized for automating fish classification diseases [3]. Considering that, this proposed to utilize transfer learning method where pre-trained models are fine tuned using domain specific dataset. To conduct this study a public dataset was collected then pre-processed. To increase model robustness adversarial training and augmented regularization were introduced. For model interpretability few Explainable AI methods (XAI) were applied.

Accurate detection and classification directly increase Bangladesh's shrimp production value through three mechanisms: First, early WSSV identification reduces mortality rates by enabling timely interventions, preserving harvest volumes. Second, precise health certification of "Healthy" class shrimp meets international export standards, commanding premium prices. Third, minimizing false negatives in BG_WSSV coinfections prevents rejected shipments, maintaining market access. By reducing disease-related losses, this system could improve product consistency. The technology's mobile compatibility further empowers farmers to optimize feeding and harvest timing based on health status, maximizing yield value per pound.

We arranged the paper using the following structure, discussion about studies in this field is presented in Section II. In section III we explain each of our methods and give detailed insights on how our research have been conducted. The outcome of our experiments has been analyzed and presented in section IV. We give a brief summary of the research and ideas about future studies in Section V.

*II. Literature Review*

Shrimp diseases pose severe economic threats to global aquaculture, driving significant research interest in automated detection systems. Deep learning (DL) has emerged as a transformative tool for early disease identification, enabling rapid, scalable solutions. This section reviews key studies advancing DL in shrimp disease classification, highlighting methodological innovations, performance gains, and limitations.

Liu et al. designed a modified YOLOv8n model using RLDD detection heads and C2f-EMCM modules to reduce computational complexity [4]. Their approach achieved 92.7% mAP@0.5 with 32.3% fewer parameters than YOLOv8n. The limitation is its reliance on bounding-box annotations for lesion detection rather than holistic classification. Compared to our work, which performs end-to-end classification without localization requirements, their method is less suitable for whole-shrimp pathology assessment.

Verma et al. proposed an AI-driven framework correlating gut microbiome dysbiosis with shrimp disease symptoms [5]. They advocated deep learning for preventive health management but provided no empirical validation of their AI component. The shortcoming is the lack of implementation details or performance metrics. Our study addresses this gap by operationalizing anatomical feature analysis through Grad-CAM++ visualizations. Tiy et al. applied probabilistic deep learning with triplet loss to identify susceptible shrimp larvae [6]. Using DenseNet121 with transfer learning, they achieved 92% accuracy. However, this work focuses exclusively on larval development stages rather than specific diseases. Our approach outperforms this with 97% accuracy for multi-disease classification in mature shrimp. Nguyen et al. employed transfer learning on CNNs to diagnose six shrimp diseases in Vietnam's Mekong Delta [7]. They attained 90.02% accuracy in field conditions but reported performance degradation with complex backgrounds. Our method overcomes this limitation through rembg background removal and FGSM adversarial training. Silva et al. developed a mobile-compatible CNN using HSV color space for industrial pigmentation grading [8]. While achieving efficient deployment, this work is not designed for disease diagnosis. We retain similar mobile compatibility while prioritizing pathological classification through advanced augmentation techniques. Rajesh et al. implemented a KNN classifier with handcrafted features to categorize shrimp health as normal/medium/abnormal [9]. Their approach achieved suboptimal accuracy due to limited feature representation. Our deep feature extraction with ConvNeXt-Tiny outperforms traditional methods by >12% accuracy. Kumar et al. created a Dense Inception CNN (DICNN) for White Spot Syndrome Virus detection [10]. They reported 97.22% accuracy but focused exclusively on WSSV. Our framework matches this accuracy while covering five additional diseases and providing explainability tools.

*III. Method and Data*

Fig. 1 represents the methodology of our research. We collected a dataset from Mendeley. The dataset was then preprocessed and split into training, validation, and test sets. Pretrained models were initially used solely for feature extraction, with our CNN model serving as the classifier. Subsequently, the pretrained models were fine-tuned and trained. Adversarial training and augmented regularization were introduced to enhance robustness.

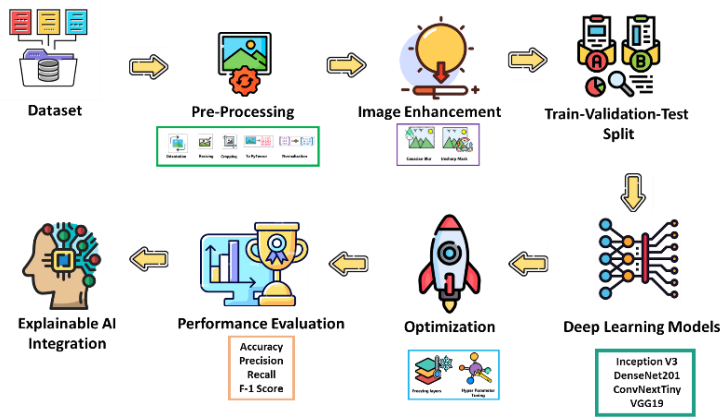

Fig. 1. Methodology

For optimization, a grid search was conducted to unfreeze layers and tune hyperparameters. After evaluating the models' performance, different variants of Grad-CAM were integrated for model interpretability..

D= pre-processing(remove background (Dataset))

$D_{train}, D_{validation}, D_{test}$=Split(D,0.7,0.15,0.15)

Pre-Trained Models= {EfficientNet, ResNet50, DenseNet, MobileNetV3, ConnNext, Xception,}

F= Feature Extract (Pre-Trained Models (D train)); Train (CNN (F, D validation))

Evaluation= Evaluate (CNN(Dtest)), Evaluate={accuracy, precision,recall,f1 score}

*A. Dataset Description and pre-processing:*

The dataset consists of 1,149 ultrasound images categorized into four classes: Healthy, Black Gill (BG), White Spot Syndrome Virus (WSSV), and a combined class exhibiting both diseases (WSSV_BG) [11]. Fig 2 illustrates sample images from the dataset.

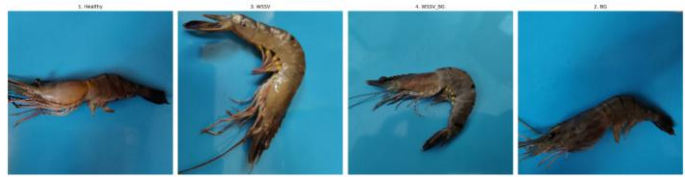

Fig. 2. Sample from Dataset

The backgrounds of the images were removed, as shown in Fig 3. All images were then pre-processed using appropriate Keras preprocessing techniques. Finally, the dataset was split into 70% for training, 15% for validation, and 15% for testing.

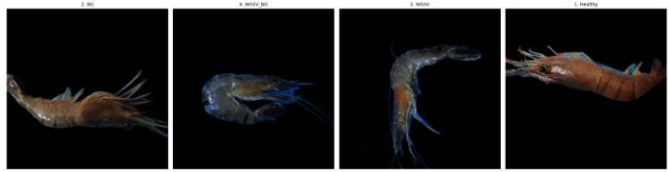

Fig. 3. Dataset after background removal

### B. Pretrained Models:

Six Convolutional Neural Network based, deep learning model were employed in this study. Initially, the pretrained model layers were frozen. Later, a grid search was used to determine the optimal number of layers to unfreeze for training each model.

*DenseNet201:*

DenseNet201 has 201 layers, and each layer is connected to every other layer in a way that information flows forward. The model includes parts like convolutional layers, dense blocks, transitional layers, global average pooling, and a fully connected layer. It uses a growth parameter to help the model learn better and prevent the vanishing gradient problem by passing features more effectively through the network

*ResNet50:*

ResNet50 is part of the Residual Network family. It includes 50 layers in total. The key parts of the model are convolutional layers, residual blocks, down sampling, and bottleneck designs. Residual blocks let the model learn the difference between the input and output, which makes training easier. There are also shortcuts that let information pass through several layers directly.

*EfficientNet:*

EfficientNet is a family of convolutional neural networks optimized for accuracy and efficiency. It uses Mobile Inverted Bottleneck Convolution (MBConv) blocks, Squeeze-and-Excitation, and a compound scaling method to uniformly scale depth, width, and resolution. This design achieves high performance with fewer parameters and lower computational cost.

*MobileNet V3:*

MobileNet is a light weight convolutional neural network. Depth-wise separable convolutions, inverted residual blocks, squeeze-and-excitation modules, neural architecture search, lightweight Attention are few of the common components of the model. Light weight attention mechanics improves the model performance greatly without adding much complexity.

*ConvNeXtTiny:*

ConvNeXt-Tiny is a smaller version of the ConvNeXt model that is efficient and still performs well. It uses special types of convolution layers, modified bottleneck structures, bigger kernel sizes, and Layer Normalization. The model follows a design inspired by transformers but has fewer parameters and less computation, making it run faster on devices with limited resources.

*MobileNet V3:*

Xception is a type of neural network that uses depthwise separable convolutions. Instead of using regular convolutions, it uses two kinds: pointwise and depthwise. This helps the network learn features from both the spatial and channel directions more efficiently. The design is based on the Inception model but is simpler because all the convolution operations are fully separable.

*CNN Model:*

We use a simple 4-layer CNN model as the top model. This model act as the classifier. The equation of this model is,

$H_1 = ReLU(G(F(X)) * W_1 + b_1)$, $H_2 = Dropout(H_1)$, $Y = H_2 * W_2 + b_2$, $Y\_hat = softmax(Y)$

Where, $F(X)$ =output of bottom model, G=global average pooling layer, H=dense layers W=weights, b=biases, Y=output

### C. Hyper Parameter tuning:

Hyperparameters significantly influence model performance and are key to optimization [12]. Batch size defines the number of samples per training step, while epochs indicate how many times the entire dataset passes through the model. The learning rate controls weight updates, often adjusted by a scheduler during training. The optimizer refines weights for better accuracy, and patience stops training early if no improvement is observed. Table I lists the hyperparameter values used in this study.

Table I. Hyper Parameter Tuning

| Optimizer | Loss Function | Scheduler | Epoch | Batch size | Patience |
|---|---|---|---|---|---|
| Adam | Sparse Categorical Crossentropy | StepLR (step_size=3) | 30 | 128 | 5 |

### D. Augmented Regularization:

Augmented Regularization can greatly improve the robustness of the model [13]. In this study we utilize both cutmix and mixup technique.

*MixUp:*

MixUp generates additional training samples by merging two images into one, as illustrated in Fig. 4. This technique blends the entire image. The equation for MixUp is as follows:

For input x and output y; $x_{new} = \lambda x_i + (1-\lambda)x_j$, $y_{new} = \lambda y_i + (1-\lambda)y_j$ where $\lambda$ is a mixup ratio drawn from a Beta distribution: $\lambda \sim Beta(\alpha, \alpha)$.

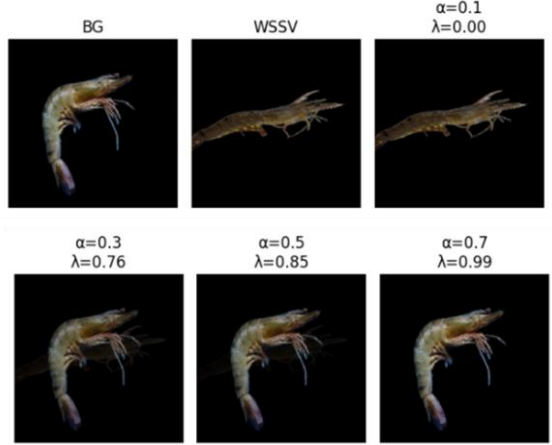

Fig. 4. MixUp image for alpha=0.2

*Cutmix:*

CutMix creates new training samples by cutting a region from one image and pasting it onto another, as shown in Fig. 4. The corresponding equation is:

For input x and output y; $x_{new} = x_i \odot M + x_j \odot (1-M)$; $y_{new} = \lambda y_i + (1-\lambda)y_j$ where M is the binary mask and $\lambda$ is the ratio of the patch area.

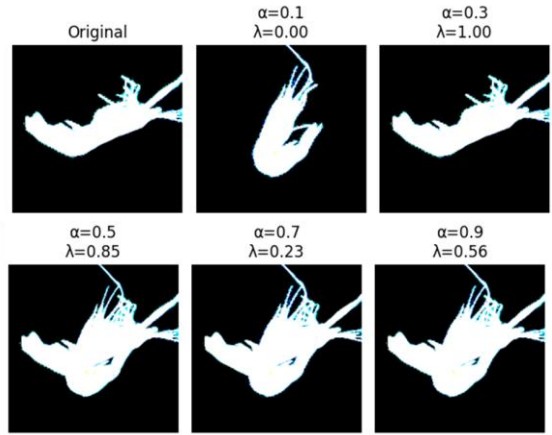

Fig. 5. CutMix image for different values

### E. Adversarial training:

Adversarial training enhances model robustness by generating and training on modified inputs that are harder to classify [14]. In this study we used Fast Gradient Sign Method. (FGSM). Adversarial examples are created using the formula: $x' = x + \epsilon \cdot sign(\nabla x\ loss(f(x), y))$, where x is the original input and $\epsilon$ controls the strength of the perturbation. Fig. 6 illustrates how the image changes with different $\epsilon$ values.

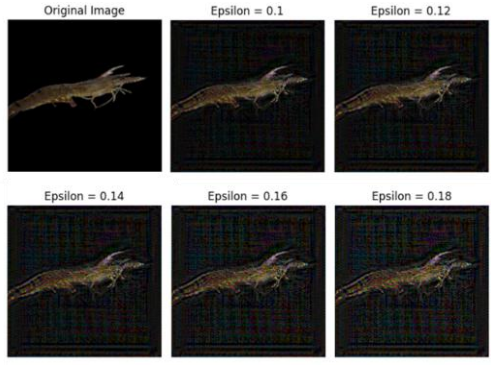

Fig. 6. Adversarial Examples created by FGSM

*F. Grad-CAM:*

Grad-CAM is a widely used technique for interpreting deep learning models by highlighting the influence of various features on the model's predictions [15]. For $k^{th}$ filter the feature map is $A^k$, $y^c$ is the score for class c, global average of gradient is $a_c^k = \frac{1}{z}\sum_{i,j} \frac{\Delta y_c}{\Delta A^k}$. Class activation map $L_{\text{Grad-CAM}} = \text{ReLU}(\sum_k a_c^k \cdot A^k)$

*IV. Result Analysis*

Table II presents a comparison of model performance on test dataset. ConvNeXt-Tiny achieved the highest accuracy at 96.88%, while MobileNet had the lowest, with only 55.21% accuracy. DenseNet, ResNet, and Xception showed similar performance, with metric scores ranging between 89% and 90%. EfficientNet performed slightly better, with scores ranging from 92% to 93% across all metrics.

Table II. Comparison Between Model for different methods

| Models | Accuracy | Precision | Recall | F1-Score |
|---|---|---|---|---|
| ConvNextTiny | 96.88 | 97.10 | 96.88 | 96.89 |
| EffecientNet | 92.71 | 93.03 | 92.71 | 92.55 |
| DenseNet201 | 90.10 | 90.23 | 90.10 | 90.07 |
| Xception | 89.06 | 89.71 | 89.06 | 88.76 |
| ResNet50 | 89.06 | 89.38 | 89.06 | 89.01 |
| MobileNetV3 | 55.21 | 39.80 | 55.21 | 44.54 |

Figure 7 shows the validation accuracy and loss for all models. ConvNeXt stands out with the highest and most stable performance during validation. Although all models achieved low validation loss, MobileNet's accuracy remained low. The other models performed well but showed some fluctuations in their validation accuracy.

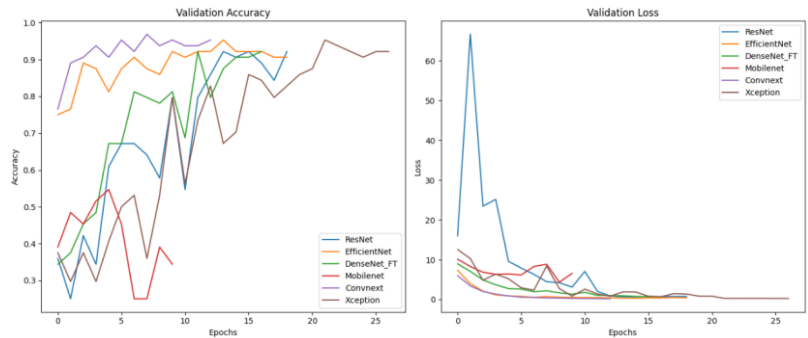

Fig. 7. Validation Accuracy and Loss for all models

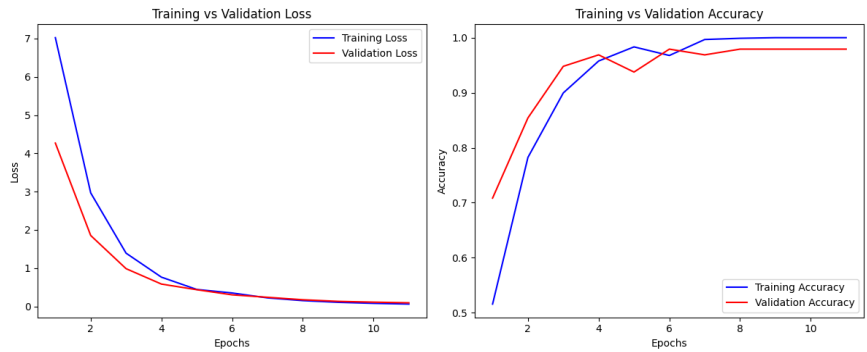

Fig. 8. Accuracy and Loss of ConvNextTiny during Training and Validation

Fig. 8 illustrates the training and validation accuracy and loss curves. The accuracy trends for both phases are closely aligned, and the loss curves nearly overlap. This indicates that the model generalizes well and is not overfitted, as there is minimal difference between training and validation performance.

Table III. Performance for Augmented Regularization

| Alpha MixUP | Alpha CutMix | Test Accuracy | Test Loss |
|---|---|---|---|
| 0 | 0 | 96.88% | 0.0430 |
| 0.2 | 0.3 | 95.42% | 0.0682 |
| 0.25 | 0.4 | 95.30% | 0.0717 |
| 0.3 | 0.5 | 93.61% | 0.1013 |
| 0.35 | 0.6 | 90.13% | 0.0967 |
| 0.4 | 0.7 | 88.95% | 0.1004 |

Table III shows the performance of the ConvNeXt model with augmented regularization using synthetic data generated via MixUp and CutMix. The model maintained strong performance, achieving over 88% accuracy in all cases. Additionally, these techniques contributed to improved robustness.

Table IV. Performance for Adversarial Training

| Epsilon Value ($\epsilon$) | Test Accuracy | Validation Loss |
|---|---|---|
| 0(original) | 96.88% | 0.093 |
| 0.1 | 94.42% | 0.104 |
| 0.12 | 92.30% | 0.109 |
| 0.14 | 88.61% | 0.124 |
| 0.16 | 79.13% | 0.117 |
| 0.18 | 70.65% | 0.581 |
| 0.2 | 59.38% | 0.676 |

Table IV presents the model's performance on various adversarial examples. As expected, the model performs best on the original images. As the value of $\epsilon$ increases, validation accuracy declines while validation loss rises. Notably, the model still achieves a strong accuracy of 88.61% and can detect distorted images even at an $\epsilon$ value of 0.14.

Table V. Classification Report for DeneseNet201

|  | Precision | Recall | F1-Score | Support |
|---|---|---|---|---|
| Healthy | 0.98 | 0.98 | 0.98 | 60 |
| BG | 1.00 | 0.94 | 0.97 | 31 |
| WSSV | 0.91 | 1.00 | 0.95 | 40 |
| WSSV_BG | 1.00 | 0.93 | 0.96 | 29 |
| Accuracy | 0.97 |  |  | 160 |
| Macro Avg | 0.97 | 0.96 | 0.97 | 160 |
| Weighted Avg | 0.97 | 0.97 | 0.97 | 160 |

Table V presents the classification report for ConvNextTiny on the test set, demonstrating strong performance across all classes. The model achieved an overall score of 98% in precision, recall, and F1-score. For the BG class, precision was perfect (100%), with recall and F1-score at 94% and 97%, respectively. The WSSV class had a perfect recall (100%), while WSSV_BG achieved perfect precision (100%). Both recall and F1-score for these two classes also exceeded 90%, indicating consistent and reliable classification.

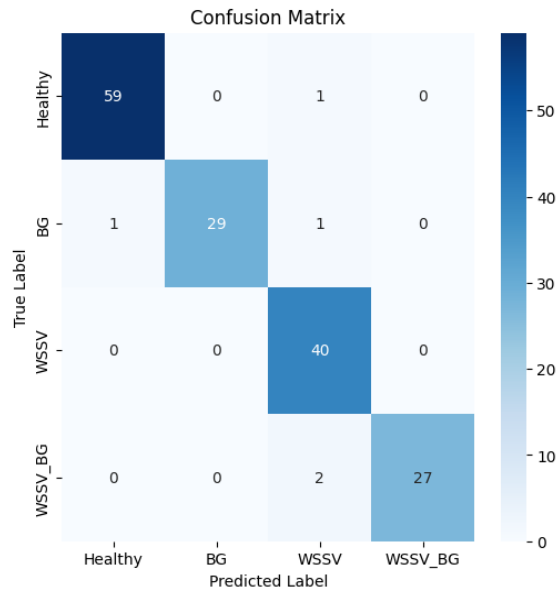
Fig. 9. Confusion Matrix

The confusion matrix shown in Fig. 8 reinforces the model's strong performance. It correctly identified all images in the WSSV class without error. There was only 1 misclassification in the Healthy class, 2 in the BG class, and 2 in the WSSV_BG class. Overall, the model made just 5 errors out of 160 images, correctly classifying 155 samples, demonstrating high accuracy and reliable predictions across all classes.

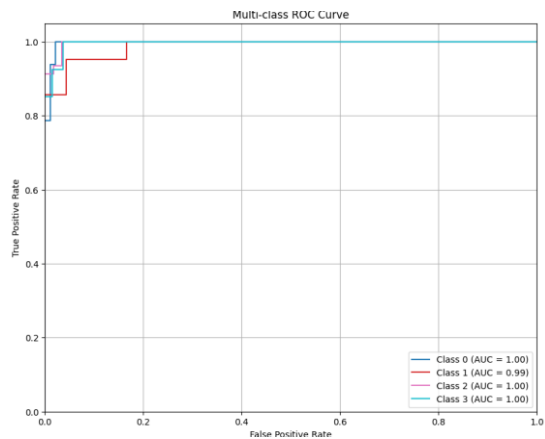
Fig. 10. Confusion Matrix

Fig. 10 shows the Receiver Operating Characteristic (ROC) curves for all classes. The Area Under the Curve (AUC) is 1.00 for three classes and 0.99 for the remaining one, indicating excellent performance. These results demonstrate that the model is highly effective at distinguishing between all classes.

Figure 11 displays heatmaps generated using different variants of Grad-CAM—Grad-CAM, Grad-CAM++, and XGrad-CAM, to interpret the model's predictions. The visualizations highlight the regions the model focuses on when making decisions. In the original image, the affected stone area is correctly marked, indicating that the model is accurately identifying disease features. As expected, areas with more intense colors represent greater importance, supporting the model's reliability in disease classification.

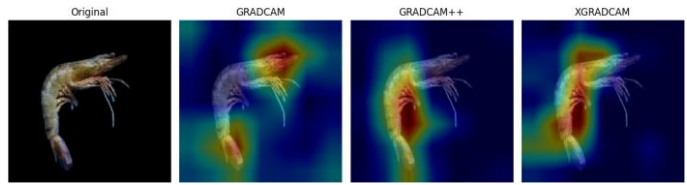
Fig. 11. Heatmap generated by different Grad-CAMs

After 1000 iterations on test set, we get standard deviation=0.0034 and mean accuracy=0.962. For 99% confidence z-score= 2.576. So, Confidence Interval, CI= 0.962 ± 2.576 * 0.0011= [0.953,0.971]

Table VI presents a comparison between our proposed models and previously published approaches. Our fine-tuned ConvNeXt model, combined with a custom CNN classifier, achieved an accuracy of 96.887%, with a 99% confidence interval of [0.953, 0.971], outperforming all prior works in this domain. Although one study reported a slightly higher accuracy of 97.22%, it did not incorporate Explainable AI techniques or apply specialized training methods to enhance model robustness, as we have done.

Table VI. Comparison with Previous works

| Ref | Models | Accuracy | Explainable AI | Robustness Specialization |
|---|---|---|---|---|
| 4 | YOLOv8 | 92.7% | No | No |
| 6 | DenseNet | 92 % | No | No |
| 7 | CNN | 90.02% | No | No |
| 9 | CNN+KNN | 88% | No | No |
| 10 | Dense Inception CNN | 97.22% | No | No |
| Proposed Model | Fine-tuned ConvNextTiny | 96.88% | Yes | Yes |

*V. Conclusion*

This study presents a comprehensive deep learning framework for automated multi-disease classification in shrimp aquaculture, offering significant improvements in accuracy, robustness, and interpretability over existing approaches. Unlike previous methods that often target single diseases or non-diagnostic features, the proposed system accurately identifies three major shrimp diseases along with the healthy stage, achieving a test accuracy of 96.88% and covering a broader range of conditions. The framework utilizes transfer learning, with model fine-tuning guided by grid search to optimize the number of trainable layers, along with hyperparameter tuning to enhance performance. Robustness is strengthened through background noise reduction, adversarial training, and augmented regularization using MixUp and CutMix techniques. To ensure interpretability, the model incorporates explainable AI (XAI) tools like Grad-CAM variants, providing visual insights into predictions and supporting actionable decisions for farmers. Designed for real-world deployment, the system is lightweight and mobile-compatible, enabling easy use in field settings without the need for advanced hardware.

While currently focused on static image analysis, future enhancements may include real-time video monitoring and integration with microbiome sensors for deeper insights into internal health. By uniting high accuracy, explainability, and practical deployment, this framework sets a new benchmark in shrimp disease management and empowers farmers with early, reliable diagnostics to reduce losses and improve productivity.